\documentclass{article}

\usepackage{microtype}
\usepackage{graphicx}
\usepackage{adjustbox}
\usepackage[table]{xcolor}
\usepackage{subfigure}
\usepackage{booktabs} 

\usepackage{threeparttable}
\usepackage{amsmath}
\usepackage{amsfonts}
\usepackage{mathtools}
\usepackage{xspace}

\usepackage{hyperref}

\usepackage[accepted]{icml2021}

\icmltitlerunning{OOD Detection Methods are Inconsistent across Datasets}

\def\shownotes{1}
\ifnum\shownotes=1
\newcommand{\authnote}[2]{[#1: #2]}
\else
\newcommand{\authnote}[2]{}
\fi
\newcommand{\mx}[1]{{\color{orange}\authnote{MX}{#1}}}
\newcommand{\ak}[1]{{\color{red}\authnote{AK}{#1}}}

\DeclareMathOperator{\sign}{sign}
\DeclareMathOperator*{\argmin}{argmin\xspace}
\DeclareMathOperator*{\argmax}{argmax\xspace}

\begin{document}

\twocolumn[
\icmltitle{No True State-of-the-Art? \\ OOD Detection Methods are Inconsistent across Datasets}




\icmlsetsymbol{equal}{*}

\begin{icmlauthorlist}
\icmlauthor{Fahim Tajwar}{cs_stanford}
\icmlauthor{Ananya Kumar}{equal,cs_stanford}
\icmlauthor{Sang Michael Xie}{equal,cs_stanford}
\icmlauthor{Percy Liang}{cs_stanford}
\end{icmlauthorlist}

\icmlaffiliation{cs_stanford}{Department of Computer Science, Stanford University, Stanford, California, USA}

\icmlcorrespondingauthor{Fahim Tajwar}{tajwar@cs.stanford.edu}

\icmlkeywords{OOD Detection, Out-of-distribution, Reliable ML}
\vskip 0.3in]



\printAffiliationsAndNotice{\icmlEqualContribution} 

\begin{abstract}
Out-of-distribution detection is an important component of reliable ML systems. Prior literature has proposed various methods (e.g.,  MSP \cite{hendrycks_17}, ODIN \cite{liang_18}, Mahalanobis \cite{lee_18}), claiming they are state-of-the-art by showing they outperform previous methods on a selected set of in-distribution (ID) and out-of-distribution (OOD) datasets. In this work, we show that none of these methods are inherently better at OOD detection than others on a standardized set of 16 (ID, OOD) pairs. We give possible explanations for these inconsistencies with simple toy datasets where whether one method outperforms another depends on the structure of the ID and OOD datasets in question. Finally, we show that a method outperforming another on a certain (ID, OOD) pair may not do so in a low-data regime. In the low-data regime, we propose a distance-based method, Pairwise OOD detection (POD), which is based on Siamese networks and improves over Mahalanobis by sidestepping the expensive covariance estimation step. Our results suggest that the OOD detection problem may be too broad, and we should consider more specific structures for leverage. 

\end{abstract}

\section{Introduction}
\label{Introduction}

When the test distribution matches the training distribution, deep neural networks (DNN) generalize well~\cite{krizhevsky_12, simonyan_15, cho_14, zhang_17}. However, when deploying a neural network in the real world, this assumption often does not hold. For example, a neural network trained on a certain set of diseases may face an unseen disease during test time, and ideally should flag such an occurrence for deferral to a human expert~\cite{mozannar_2020, rajpurkar_20}.
This work focuses on OOD detection~\cite{hendrycks_17, liang_18, lee_18}, the binary classification task of detecting whether a test example is sampled from a \emph{known} distribution $\mathcal{D}_{in}$ (an in-distribution (ID) example) or an \emph{unknown} distribution $\mathcal{D}_{out}$ (an out-of-distribution (OOD) example).

\begin{table} \label{table:comparisonWithoutOOD}
    \centering
    \begin{adjustbox}{width=\columnwidth,height=0.5in,center}
        \begin{tabular}{ccccccccc}
            \toprule
             & \multicolumn{2}{c}{MSP} & \multicolumn{2}{c}{ODIN} & \multicolumn{2}{c}{Mahalanobis} & \multicolumn{2}{c}{POD} \\
\cmidrule(){2-9}
 & Full &  Low  & Full & Low  &  Full & Low & Full & Low  \\
\midrule
MSP &  ---  & ---   &   2/16 &    3/16  &    2/16 &   9/16 &   3/16 &   3/16 \\
ODIN &   14/16 &   13/16 &  --- & ---  &   4/16 &   13/16 &   4/16 &   13/16 \\
Mahalanobis &   14/16 &   7/16  &   12/16 &   3/16 & ---  & ---  &   12/16 &   6/16 \\
POD &   13/16 &   13/16 &   12/16 &   3/16 &   4/16 &   10/16 & ---  & ---  \\
\bottomrule
        \end{tabular}
    \end{adjustbox}
    \caption{Comparison table showing the number of ID-OOD dataset pairs (out of 16) on which the row method outperforms the column method on OOD detection AUROC, without using any OOD samples for training/hyper-parameter tuning. No method does consistently better than any other method across the 16 pairs (none of the numbers are 0 or 16). Comparisons are also sensitive to in-distribution training dataset sizes --- for example, Mahalanobis (third row) does better than ODIN on 12/16 pairs using the full ID train set, but on only 3/16 pairs in the low data regime.}
\end{table}

In this work, we consider two popular groups of OOD detection methods: predictive score-based (e.g., MSP~\cite{hendrycks_17}, ODIN~\cite{liang_18}) and distance-based (e.g., Mahalanobis~\cite{lee_18}) methods. Prior literature shows empirically that their methods do better than other commonly used baselines on a number of ($\mathcal{D}_{in}$, $\mathcal{D}_{out}$) pairs. When we standardize training details and test on a larger number of datasets, we find that no single method consistently outperforms others across different ($\mathcal{D}_{in}$, $\mathcal{D}_{out}$) pairs on the OOD detection task.

Specifically, under the assumption that no OOD data is seen during training/hyper-parameter tuning (similar to~\citet{hsu_20}), we run experiments on a set of 3 ID datasets (CIFAR-10, CIFAR-100 and SVHN) and 7 OOD datasets (CIFAR-10, CIFAR-100, SVHN, CelebA, STL-10, TinyImageNet, LSUN) with a standardized training procedure and show that none of three popular OOD detection methods (MSP, ODIN, Mahalanobis) consistently does better than the others (see Table \ref{table:comparisonWithoutOOD}) on the $16$ different ($\mathcal{D}_{in}$, $\mathcal{D}_{out}$) pairs in terms of OOD detection AUROC. Comparisons are also sensitive to in-distribution training dataset sizes --- for example, Mahalanobis does better than ODIN on 12 out of 16 cases with the entire ID train dataset; but on only 3 out of 16 cases when $10\%$ of the ID train data is used, so there is no clear winner even if we average the metrics across datasets.

To examine possible reasons why these inconsistencies can arise, and show that these issues are fundamental and not specific to implementation details, we provide toy datasets that demonstrate whether predictive score-based methods do better or worse than distance-based methods depends on the particular ($\mathcal{D}_{in}$, $\mathcal{D}_{out}$) pair, and neither type of method is universally better. Specifically, both a linear classifier and a distance based classifier achieve $100\%$ ID test classification accuracy in the two toy datasets in $\mathbb{R}^2$, but ---

\begin{itemize}
    \item On toy dataset 1 (Figure \ref{fig:toyDataset1}), a distance-based method achieves $100\%$ AUROC on the OOD detection task, whereas MSP (based on linear classifier) achieves $0\%$.
    \item In contrast, on toy dataset 2 (Figure \ref{fig:toyDataset2}), a distance-based method achieves $32.5\%$ AUROC in contrast to MSP's $100\%$ AUROC on the OOD detection task.
\end{itemize}

Given that inconsistencies are fundamental in the broad OOD detection setting, we consider more specific settings where differences between methods may be more consistent. For example, with access to some OOD data for hyper-parameter tuning, we see more consistent trends for OOD detection methods. In the low-data regime, we propose a distance-based method, Pairwise OOD detection (POD), that is based on Siamese networks~\cite{koch_15} and improves the Mahalanobis method~\cite{lee_18} by crucially sidestepping the expensive covariance matrix estimation step. POD improves over MSP and Mahalanobis in 13 and 10 out of 16 dataset pairs respectively in the low data regime (Table \ref{table:comparisonWithoutOOD}).

Our contributions can be summarized as: 
\begin{enumerate}
    \item We run three popular baselines (MSP, ODIN, Mahalanobis) and POD
    on a standard set of 16 ($\mathcal{D}_{in}$, $\mathcal{D}_{out}$) pairs, and show that no method performs better than the others consistently on the full data setting.
    \item We present toy datasets in $\mathbb{R}^2$ and show that none of the two popular classes of OOD detection methods (predictive score and distance-based methods) do better than the other consistently, and their performance depends on the particular ($\mathcal{D}_{in}$, $\mathcal{D}_{out}$) pair.
    \item Finally, we show that this inconsistency is even more pronounced in the low-data setting, i.e., methods that tend to do better than another in the high data setting may do worse when the ID training dataset size has decreased, so even if we average metrics across datasets there is no clear best method.
\end{enumerate}


Given the inconsistencies in the broad OOD detection problem setting, we believe more specific settings (e.g.,  some access to OOD samples for hyper-parameter tuning, low data regime) are necessary for targeted improvements.

\section{Problem Description} \label{section:problemDescription}

Here we formally define the OOD detection problem following prior works such as~\citet{liang_18}. Let $\mathcal{X}$ be the input space, $\mathcal{Y}$ be the label space, and $\mathcal{D}_{in}$ be a known distribution over $\mathcal{X} \times \mathcal{Y}$ with $C$ classes. We have access to a set of $N$ examples, $D_{in}^{train} = \{(x_i, y_i)\}_{i=1}^N$ sampled from $\mathcal{D}_{in}$ for training. Consider any $x \in \mathcal{X}$, a model trained on $D_{in}^{train}$ will output a score $S(x) \in \mathbb{R}$, where $S(x)$ should be higher for in-distribution $x$. Then we can use $S(x)$ as a heuristic for OOD detection, i.e., we pick a threshold $T$, and say $x$ is ID if $S(x) \geq T$, or otherwise $x$ is OOD. 

During \emph{test time}, we have examples $D_{out}^{test}$ drawn from an unseen distribution $\mathcal{D}_{out}$, where the set of classes are disjoint from the set of classes in $\mathcal{D}_{in}$. We also have a set of examples drawn from $\mathcal{D}_{in}$ that are not in $D_{in}^{train}$, we denote this as $D_{in}^{test}$. Then OOD detection becomes a binary classification problem where examples from $D_{in}^{test}$ are ID or positive, and examples from $D_{out}^{test}$ are OOD or negative. We use two popular metrics ~\cite{hendrycks_19} to measure the performance of an OOD detection method:

\begin{enumerate}
    \item \textbf{AUROC}: Area under the receiver operating characteristic curve, which is the plot of true positive rate vs false positive rate for a binary classification problem.
    \item \textbf{FNR@95}: False negative rate (fraction of ID examples misclassified as OOD) when 95\% of OOD examples are classified as OOD (true negative). Note that~\citet{hendrycks_19} denotes this as FPR@95, since they denote OOD examples as positive instead of negative and vice versa. 
\end{enumerate}

\section{Methods} \label{subsection:methods}

We compare four OOD detection methods (3 popular baselines and our method) in this work. 

\begin{enumerate}
    \item \textbf{MSP} \cite{hendrycks_17}: Maximum softmax probability (MSP) is a commonly used predictive score based method. We first train a neural network $f$ to classify the $C$ in-distribution classes. Given input $x$, let $f^i(x)$ denote the model's confidence that $x$ has label $i$. The temperature-scaled softmax score for class $i$ is:
    \begin{equation} \label{equation:S(x;T)}
        S_{soft}^i(x; T) = \frac{\exp{(f^i(x) / T)}}{\sum_{j = 1}^C \exp{(f^j (x) / T)}}
    \end{equation}

    The MSP score for $x$ is simply the maximum probability across the classes:
    
    \begin{equation} \label{equation:MSP}
        S_{MSP}(x) = \max_{i \in \{1, \dots, C\}} S_{soft}^i(x; T=1)
    \end{equation}
    
    \item \textbf{ODIN} \cite{liang_18}: ODIN builds on MSP by temperature scaling with some $T > 0$ and pre-processes the inputs with noise $\epsilon > 0$ to improve OOD detection performance. Let $\hat{x} = G_{ODIN}(x; \epsilon, T)$ be the pre-processed version of $x \in \mathcal{X}$ (Appendix \ref{appendix:odinParameters}). Then the ODIN score for $x$ is
    
    \begin{equation} \label{equation:ODIN}
        S_{ODIN}(x; T) = \max_{i \in \{1, \dots, C\}} S_{soft}^i(\hat{x} ; T)
    \end{equation}
    
    \item \textbf{Mahalanobis} \cite{lee_18}: Mahalanobis is a popular distance based method. Let $h(x)$, $\mu_i$ and $\Sigma$ be the neural network representation, class mean for label $i$, and covariance matrix at the neural network’s penultimate layer respectively. Let $\hat{x} = G_{Maha}(x ; \epsilon)$ be the pre-processed version of input $x$. For details on how $\mu_i$, $\Sigma$ and $\hat{x}$ are calculated, see Appendix \ref{appendix:mahalanobisParameters}. We define
    \begin{equation} \label{equation:Maha_k_layer}
        \begin{split}
            S_{Maha}(x) = \max_{i=\{1,\dots,C\}} \{ - (h(\hat{x}) - \mu_i)^T \Sigma^{-1} \\
            (h(\hat{x}) - \mu_i) \}
        \end{split}
    \end{equation}
    
    Note that~\citet{lee_18} use the weighted sum of the Mahalanobis distance from each layer of the neural network. However, the weights are tuned using OOD samples. Following more recent work~\cite{liu_20}, we use the Mahalanobis distance in only the penultimate layer.
    
    \item \textbf{Pairwise OOD Detection (POD)} (Ours):  We propose a distance-based method for the low-data regime -- “Pairwise OOD Detection” (POD). Let $h(x) \in \mathbb{R}^d$ be the neural network's penultimate layer’s representation for test sample $x$. Suppose we have $M$ images from each of the $C$ ID classes i.e., $\{\{z_{ij}\}_{j=1}^M\}_{i=1}^C$ where $z_{ij}$ is the $j$-th example from the $i$-th ID class. We define
    
    \begin{equation} \label{equation:pairwise_class_i}
        S_{POD}^i(x) = \frac{1}{M} \sum_{j = 1}^M ||h(x) - h(z_{ij})||_2^2
    \end{equation}
    
    \begin{equation} \label{equation:POD}
        S_{POD}(x) = - \min_{i \in \{1, \cdots, C\}} S_{POD}^i(x)
    \end{equation}
    
    In Appendix \ref{appendix:accumulators}, we provide justification for our choice of $S_{POD}$.
\end{enumerate}

Among these methods, ODIN and Mahalanobis require extra hyper-parameters to be tuned using OOD samples. We do not typically have access to $\mathcal{D}_{out}$, and methods such as Outlier exposure ~\cite{hendrycks_19} or Energy based fine-tuning ~\cite{liu_20} do better when there is some access to OOD samples. We record results with standardized hyper-parameters in Table \ref{table:comparisonWithoutOOD}, and OOD sample tuned hyper-parameters in Table \ref{table:comparisonWithOOD}. Please see Appendix \ref{appendix:additionalHyperparameterTuning} for details on OOD sample dependent hyper-parameters.

\section{OOD Detection methods are inconsistent}

\subsection{Toy datasets}

Using toy datasets, we demonstrate that an OOD detection method's relative performance depends on the ($\mathcal{D}_{in}$, $\mathcal{D}_{out}$) pair in question. Specifically, we present two toy datasets in $\mathbb{R}^2$ (see Appendix \ref{appendix:toyDatasets} on how to generate them). For the predictive score based model, we consider MSP on top of a linear classifier trained on $D_{in}^{train}$ dataset. For the distance based method, we use the negative of the minimum distance from the test sample $x$ to any sample in $D_{in}^{train}$ for ID classification or OOD detection task.
 
\begin{figure*}[ht]
    \centering
    \begin{minipage}[b]{0.40\linewidth}
         \centering
         \includegraphics[width=0.8\columnwidth,    height=0.6\columnwidth]{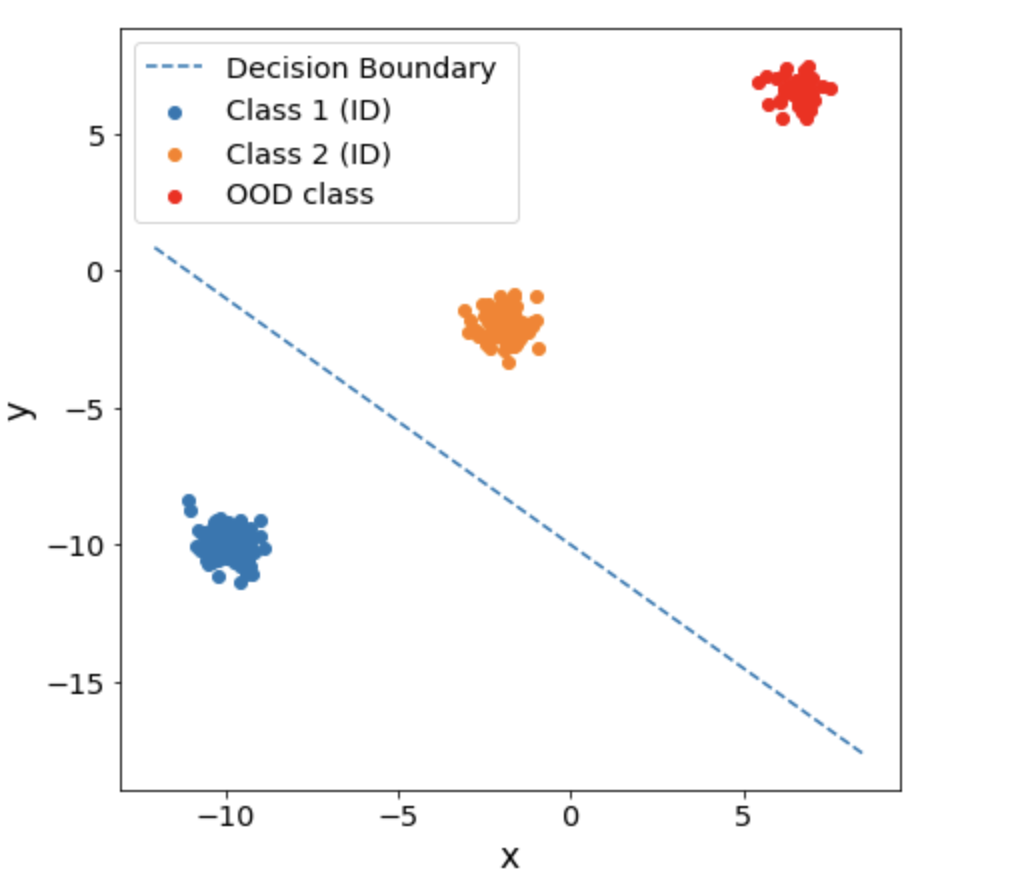}
         \caption{Toy Dataset 1. Here, distance-based methods do better than MSP due to the OOD examples being far away from the decision boundary between classes, and thus fooling MSP into giving them a higher confidence than ID samples.}
         \label{fig:toyDataset1}
    \end{minipage}
    \quad
    \begin{minipage}[b]{0.40\linewidth}
    \centering
         \includegraphics[width=0.8\columnwidth,    height=0.6\columnwidth]{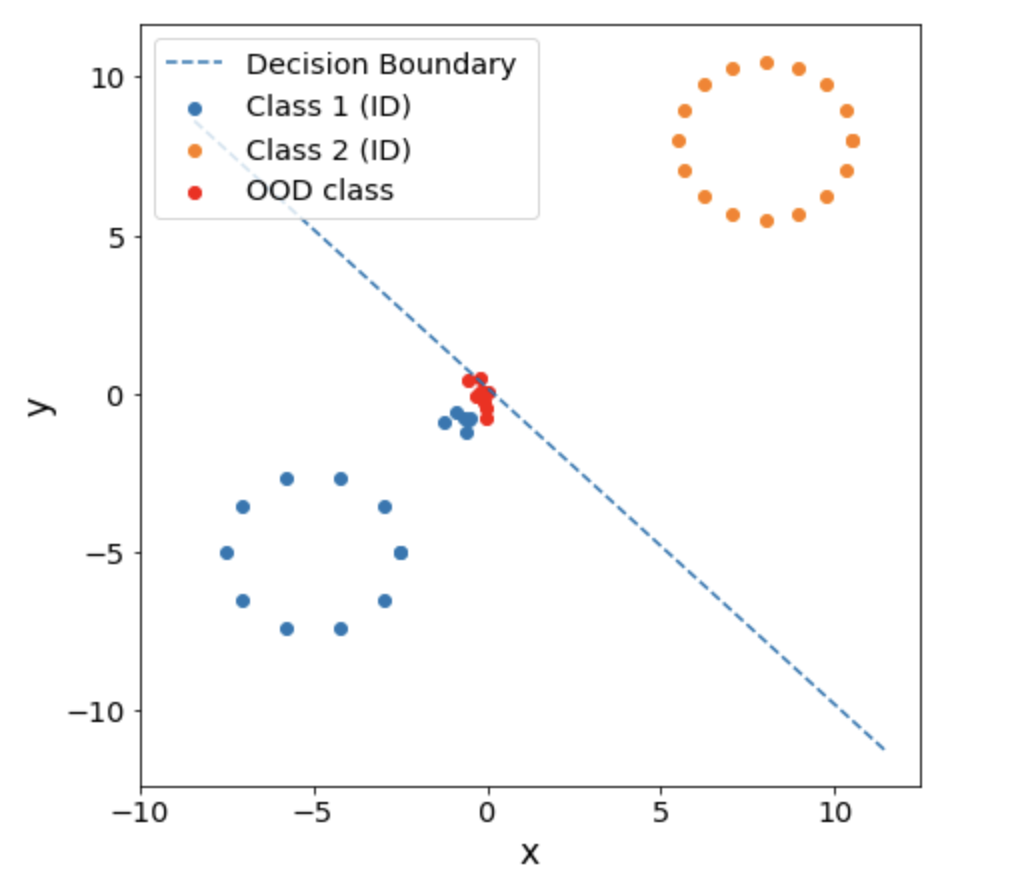}
         \caption{Toy Dataset 2. Here, MSP does better than distance based methods since OOD examples are very close to the decision boundary between classes and so have lower confidence scores, whereas ID outliers close to the OOD examples fool the distance based method.} 
         \label{fig:toyDataset2}
    \end{minipage}
\end{figure*}

Now on both of these datasets, the predictive score and distance-based methods achieve $100\%$ accuracy on $D_{in}^{test}$. However, on toy dataset 1 (Figure \ref{fig:toyDataset1}), the OOD examples are far away from the decision boundary and any ID examples. This makes the MSP predict higher confidence scores on the OOD examples than the ID examples, and so MSP achieves $0\%$ AUROC on the OOD detection task. However, for the same reason the distance-based method achieves $100\%$ AUROC. In contrast, on toy dataset 2 (Figure \ref{fig:toyDataset2}), the two ID classes are linearly separable and the majority of the ID samples are spread out. The OOD examples are closer to the decision boundary than ID examples, and there are ID outliers very close to the OOD examples as well. Due to OOD examples being closer to the decision boundary, MSP achieves $100\%$ AUROC. However, the distance based method is fooled by the ID outliers, and achieves only $32.5\%$ AUROC. We note that~\citet{liu_jeremiah_20} describes 2D classification benchmarks similar to toy dataset 1 where deep neural networks have very low uncertainty on OOD data and hence fail to classify them.

\subsection{Standard/Full data OOD detection setting} \label{subsection:fullDataSetting}

We demonstrate the inconsistency among popular OOD detection methods by designing a standardized set of experiments (Appendix \ref{appendix:expSetup}). We use the full ID train dataset, and have no access to any OOD sample for training/tuning. 

\textbf{Datasets:} We use 3 ID datasets (CIFAR-10, CIFAR-100 and SVHN) and 7 OOD datasets (CIFAR-10, CIFAR-100, SVHN, CelebA, STL-10, TinyImageNet and LSUN) to form 16 different ($\mathcal{D}_{in}$, $\mathcal{D}_{out}$) pairs. See Appendix \ref{appendix:dataPairConstruction} for (ID, OOD) pairs that are excluded because of class overlaps.
    
\textbf{Model Architecture:} We use ResNet-34 \cite{he_16} as the network backbone for all of our experiments. 

The inconsistency across different methods is shown in Table \ref{table:comparisonWithoutOOD}.

\subsection{Low data OOD detection setting} \label{subsection:lowDataSetting}
To investigate whether there is more consistency in specific settings, we do the same experiments in a low-data regime, where we use only 10\% of each ID train dataset $D_{in}^{train}$. In the low-data regime (Table \ref{table:comparisonWithoutOOD}), ODIN does better than each of the other methods on $13/16$ (ID, OOD) pairs, whereas in the full-data regime ODIN only did better than Mahalanobis and POD on $4/16$ pairs. In contrast, Mahalanobis performs much worse in the low data regime, which we attribute to the statistically expensive covariance matrix estimation step. Finally, POD outperforms Mahalanobis in the low data setting, possibly by not requiring covariance estimation. This reveals that methods' performance depends on not only the (ID, OOD) pair but also the ID dataset size.

\subsection{With Access to OOD Samples}

We further specialize the setting by doing the experiments in Sections \ref{subsection:fullDataSetting} and \ref{subsection:lowDataSetting} while letting ODIN and Mahalanobis use OOD samples to tune hyper-parameters (Appendix \ref{appendix:additionalHyperparameterTuning}). Table \ref{table:comparisonWithOOD} shows that ODIN and Mahalanobis improve in the full data setting when using OOD examples. However, significant inconsistency remains between ODIN and Mahalanobis themselves, and specially in the full vs low data regime, where their performance order switches.

\begin{table} \label{table:comparisonWithOOD}
    \centering
    \begin{adjustbox}{width=\columnwidth,height=0.5in,center}
        \begin{tabular}{ccccccccc}
            \toprule
             & \multicolumn{2}{c}{MSP} & \multicolumn{2}{c}{ODIN} & \multicolumn{2}{c}{Mahalanobis} & \multicolumn{2}{c}{POD} \\
\cmidrule(){2-9}
 & Full &  Low  & Full & Low  &  Full & Low & Full & Low  \\
\midrule
MSP & ---  & ---  &      2/16   &       0/16 &       2/16 &       6/16  &       3/16  &       3/16 \\
ODIN  &     14/16  &      16/16  &  ---  &  ---  &       5/16  &     10/16         &      8/16 &      16/16 \\
Mahalanobis  &     14/16  &      10/16  &     11/16       &      6/16    &  ---  & ---  &      13/16    &      10/16 \\
POD  &      13/16  &     13/16 &       8/16       &       0/16 &      3/16  &      6/16          &   ---      &    ---      \\
\bottomrule

        \end{tabular}
    \end{adjustbox}
    \caption{Comparison table showing how many ID-OOD dataset pairs (out of 16) the row method outperforms the column method on OOD detection AUROC. In contrast to Table \ref{table:comparisonWithoutOOD}, \comment{The key difference between this table and Table \ref{table:comparisonWithoutOOD} is that the latter assumes no OOD samples are available, but }we let ODIN and Mahalanobis use OOD samples for hyper-parameter tuning in this table.}
\end{table}

\section{Discussion and Conclusion}

\comment{Our fundamental result is that the current setting of OOD detection (\ref{section:problemDescription}) is too broad. Some prior works have also discussed this issue --- ~\citet{ahmed_2020} considers detecting semantic anomalies within the same distribution \comment{(e.g., withholding an ID class as anomalous)}instead of samples from a different distribution as a more practical OOD detection problem\ak{who's more practical? Ahmed and Courville? Or us};~\citet{ruff_2020} discusses the same issue and demonstrates the lack of consistency among anomaly detection methods\ak{What's the difference between anomaly detection and OOD detection}\mx{Consider replacing with: \citet{ahmed_2020} and~\citet{ruff_2020} consider detecting low probability examples from the training distribution $\mathcal{D}_{in}$ instead of OOD examples from a different distribution and find a lack of consistency among these in-distribution anomaly detection methods}; ~\citet{hsu_20} and~\citet{huang_21} noticed performance inconsistency between Mahalanobis and MSP for semantic shifts and large semantic spaces\ak{What's something we contribute they don't?}. We suggest further studies into the possible correlation between dataset statistics and the rank of performance of an OOD detection method, and conclude that the community might benefit from moving away from the practice of coming up with a method and testing it on certain (ID, OOD) pairs, and look for more specific problem settings. 

Our fundamental result is that the current setting of OOD detection is too broad. Some prior works have also discussed this issue and proposed different settings. For example, the ID and OOD distributions can be very different (e.g., CIFAR-10 vs SVHN) in our setting;~\citet{ahmed_2020} argue that only the case when OOD examples are semantically similar to ID ones is of practical importance, and propose benchmarks to withhold one ID class during training and detecting examples from the held out class during testing as OOD.~\citet{ruff_2020} consider detecting low probability examples (e.g., blurred images, objects with defects) from the training distribution instead of OOD examples from an unseen distribution and find a lack of consistency among these in-distribution anomaly detection methods. Some failure modes of Mahalanobis method have also been discovered ---~\citet{hsu_20} construct a dataset containing semantic and non-semantic shifts based on DomainNet~\cite{peng_19} and show that Mahalanobis performs worse than MSP baseline. \citet{ren_21} show the same for semantically similar distributions (e.g., CIFAR-10 vs CIFAR-100), and~\citet{huang_21} for large semantic spaces (e.g., ImageNet). We suggest further studies into settings where certain OOD detection methods perform well, and conclude that the community might benefit from moving away from the practice of coming up with a method and testing it on certain (ID, OOD) pairs, and instead look for more specific problem settings.}

Our fundamental result is that the current setting of OOD detection is too broad. Some prior works have also discussed this issue and proposed alternative settings. For example,~\citet{ahmed_2020} proposed benchmarks where they withhold one ID class during training and want to detect examples from the withheld class as OOD during testing, instead of a possibly very different unseen distribution (e.g., CIFAR-10 vs SVHN). Similarly,~\citet{ruff_2020} considers detecting unseen corruptions of images from seen classes as the meaningful anomaly detection problem, and shows inconsistency among the performance orders of various anomaly detection methods in this modified problem setting. Prior literature have also found failure modes for the popular Mahalanobis method --- for example,~\citet{ren_21} discusses that MSP baseline outperforms Mahalanobis when ID and OOD distributions are very similar. For details on these alternate OOD detection settings and failure modes, please see Appendix~\ref{appendix:relatedWorks}.

We suggest further studies into the relationship between dataset statistics and how an individual OOD detection method will perform, and conclude that the community might benefit from moving away from the practice of coming up with a method and testing it on certain (ID, OOD) pairs, and instead look for more specific problem settings where OOD detection methods will perform consistently.

\newpage

\section*{Acknowledgements}

We thank Chelsea Finn, Tengyu Ma, Yining Chen, Tom Knowles, and Dan Hendrycks for giving us comments and suggestions during the course of this work, Weitang Liu for his clarifications on CelebA dataset transformations~\cite{liu_20}, and anonymous reviewers for their suggestions on how to improve the paper. We would also like to thank Sumaita Sadia Rahman for her help with proof-reading the entire paper. This work was supported by NSF Award Grant No. 1805310. Fahim Tajwar was supported by Stanford VPUE (Vice Provost for Undergraduate Education) via the Computer Science department's CURIS program, Ananya Kumar was supported by a Stanford Graduate Fellowship, and Sang Michael Xie was supported by an NDSEG fellowship.

\section*{Reproducibility}

We have deposited all our code and instructions on how to replicate the results in this paper in this  \href{https://github.com/tajwarfahim/OOD_Detection_Inconsistency}{GitHub repository}.

\nocite{zagoruyko_16}
\nocite{he_16}
\nocite{krizhevsky_12}
\nocite{simonyan_15}
\nocite{cho_14}

\nocite{zhang_17}
\nocite{mozannar_2020}
\nocite{rajpurkar_20}
\nocite{nguyen_15}
\nocite{hendrycks_17}

\nocite{szegedy_14}
\nocite{moosavi_17}
\nocite{liang_18}
\nocite{lee_18}
\nocite{amodei_16_concrete}

\nocite{goodfellow_15}
\nocite{hoye_21}
\nocite{prabhu_18}
\nocite{prabhu_19_few_shot_learning}
\nocite{prabhu_19_open_set_medical_diagnosis}

\nocite{shi_21}
\nocite{hendrycks_19}
\nocite{liu_20}
\nocite{hsu_20}
\nocite{koch_15}

\nocite{chen_20}
\nocite{reed_21}
\nocite{netzer_11}
\nocite{krizhevsky_09}
\nocite{liu_15}

\nocite{coates_11}
\nocite{deng_09}
\nocite{yu_15}
\nocite{loshchilov_17}
\nocite{huang_21}

\nocite{lecun_10}
\nocite{ruff_2020}
\nocite{ahmed_2020}
\nocite{tinyimagenet}
\nocite{liu_jeremiah_20}

\bibliography{ood_detection}
\bibliographystyle{icml2021}

\newpage

\appendix

\section{Related works} \label{appendix:relatedWorks}

Prior literature has discussed the issue of OOD Detection problem setting being too broad and as a response proposed alternate settings and benchmarks. Several works also discussed specific failure modes of OOD detection methods such as Mahalanobis~\cite{lee_18}. 

\textbf{Alternate OOD detection settings:} In our problem setup (Section~\ref{section:problemDescription}) similar to prior works such as~\citet{liang_18}, there is no assumption for the ID distribution $\mathcal{D}_{in}$ and OOD distribution $\mathcal{D}_{out}$, except that their underlying classes in the label space $\mathcal{Y}$ have to be different. This implies that the ID and OOD distributions in question can be very different (e.g., CIFAR-10 vs SVHN) but also very similar (e.g., CIFAR-10 vs CIFAR-100).~\citet{ahmed_2020} argue that only the case when OOD examples are semantically similar to ID ones is of practical importance. They propose benchmarks where during training they withhold one ID class and during testing they consider examples from the held out class as OOD. However,~\citet{ahmed_2020} only considers two OOD detection methods (MSP and ODIN), and do not notice the inconsistency between their performance with only 3 datasets (CIFAR-10, STL-10, and ImageNet).

\citet{ruff_2020} describe the anomaly detection problem as detecting low probability examples (e.g., blurred images, objects with defects) from the classes in the training distribution instead of examples from different classes of an unseen distribution. For their problem setting, the outliers are images with corruption or defects from the training classes but the corruptions or defects themselves are not seen during training. In this problem setting,~\citet{ruff_2020} compares 9 anomaly detection methods and notices inconsistency between their performance orders on two datasets.

\textbf{Failure modes of OOD detection methods:} Prior literature have discovered particular cases where Mahalanobis~\cite{liang_18}, a popular OOD detection method, is outperformed by the simple MSP~\cite{hendrycks_17} baseline.

\citet{hsu_20} describes two types of out-of-distribution data, namely those generated from semantic and non-semantic shifts. Here semantic shift is defined as OOD data differing from ID data in the meaning of the classes, and non-semantic shift is defined as ID and OOD data differing in image types but representing the same classes, i.e., real images of an aeroplane being ID and hand-drawn sketches of an aeroplane being OOD data.~\citet{hsu_20} construct datasets containing semantic and non-semantic shifts based on the DomainNet~\cite{peng_19} dataset, and show that in most cases Mahalanobis is outperformed by the MSP baseline, except when to detect random noise.~\citet{ren_21} shows the same for semantically similar distributions (e.g., CIFAR-10 vs CIFAR-100), and~\citet{huang_21} for large semantic spaces (e.g., ImageNet).

\section{Experiment setup} \label{appendix:expSetup}

\subsection{In-distribution (ID) datasets} \label{appendix:idDatasets}
We use three commonly used In-distribution datasets in our experiments:

\begin{enumerate}
    \item \textbf{CIFAR-10} \cite{krizhevsky_09}: CIFAR-10 contains 50,000 train and 10,000 test images, separated into 10 disjoint classes. All images are of size 32 x 32, and have three channels. The image classes are similar but disjoint from those of CIFAR-100.
    
    \item \textbf{CIFAR-100} \cite{krizhevsky_09}: CIFAR-100 contains 50,000 train and 10,000 test images of size 32 x 32, similar to CIFAR-10. However, the dataset has 100 classes, which can be grouped into 20 super-classes. Each super-class in CIFAR-100 containing 5 finer-grained classes. The classes in CIFAR-10 and CIFAR-100 are mutually disjoint.
    
    \item \textbf{SVHN} \cite{netzer_11}: SVHN contains 73,257 train and 26032 test images, each with size 32 x 32. The original dataset contains some extra train examples which we do not use in our experiments. There are 10 classes in this dataset representing the 10 different digits in English.
\end{enumerate}

For all ID datasets, we use the standard train-test split, and refer to them as $\mathcal{D}_{in}^{train}$ and $\mathcal{D}_{in}^{test}$, respectively.

\subsection{Out-of-distribution (OOD) datasets} \label{appendix:oodDatasets}

We use all of the three (CIFAR-10, CIFAR-100 and SVHN) datasets mentioned in \ref{appendix:idDatasets} as OOD datasets. Furthermore, we use the following additional OOD datasets:

\begin{enumerate}
    \item \textbf{CelebA} \cite{liu_15}: This dataset contains images of celebrities' faces. We resize the CelebA images into 32 x 32, and use the standard test split, which contains 19,962 images.
    
    \item \textbf{STL-10} \cite{coates_11}: This dataset contains almost the same classes as CIFAR-10, but the image sizes are larger. We resize the STL-10 images into 32 x 32, and use the standard test split, which contains 8000 images.
    
    \item \textbf{TinyImageNet} \cite{tinyimagenet, deng_09, liang_18}: TinyImageNet contains a subset of the ImageNet dataset \cite{deng_09} which has 10,000 test images divided into 200 different classes.~\citet{liang_18} further describes two datasets created by randomly cropping and resizing the test images to size 32 x 32 respectively. We use the resized dataset in our work here. Descriptions of this dataset (with links to download it as well) are given here: \url{https://github.com/ShiyuLiang/odin-pytorch}
    
    \item \textbf{LSUN} \cite{yu_15, liang_18}: The LSUN dataset, also known as the \textbf{L}arge-scale \textbf{S}cene \textbf{UN}derstanding dataset, has 10,000 test images divided into 10 different classes. Similar to the TinyImageNet dataset,~\citet{liang_18} constructs two datasets by randomly cropping and resizing the test images into size 32 x 32. We use the resized dataset in our work here. Further details, as well as intruction on how to download the dataset, can be obtained here: \url{https://github.com/ShiyuLiang/odin-pytorch}
\end{enumerate}

For all OOD datasets where there is a standard train-test split (CIFAR-10, CIFAR-100, SVHN, CelebA, STL-10) we use only the test split as OOD samples, and refer to it as $\mathcal{D}_{out}^{test}$. For the datasets where such a split is not available (TinyImageNet and LSUN), we use the entire dataset as $\mathcal{D}_{out}^{test}$.

\subsection{Constructing the set of (ID, OOD) pairs} \label{appendix:dataPairConstruction}

We construct 16 total (ID, OOD) pairs from the 3 ID datasets in \ref{appendix:idDatasets} and 7 OOD datasets in \ref{appendix:oodDatasets}. One may notice that we can form 21 different (ID, OOD) pairs; however, (CIFAR-10, CIFAR-10), (CIFAR-100, CIFAR-100) and (SVHN, SVHN) are not valid pairs.

Furthermore, we do not consider ($\mathcal{D}_{in}$, $\mathcal{D}_{out}$) pairs that share significant number of common classes and were not used in earlier literature, for example (CIFAR-10, STL-10) and (CIFAR-100, CelebA). CIFAR-10 and STL-10 share 9 out of 10 classes among them, and CIFAR-100 contains a "people" super-class containing facial images similar to CelebA. Hence we exclude (CIFAR-10, STL-10) and (CIFAR-100, CelebA), and arrive at 16 (ID, OOD) pairs for our experiments. Note that (CIFAR-10, TinyImageNet) and (CIFAR-100, TinyImageNet) may have some pretty similar classes; however the number of classes shared is pretty low and earlier literature~\cite{lee_18, liang_18} uses these pairs, so we decided to use them in our work as well.

\subsection{Model architecture and training details} \label{appendix:modelArchTranining}

\begin{table*}[t]
  \centering
  \begin{adjustbox}{width=\textwidth,center}
  \begin{tabular}{ccccccc}
     \toprule
     Training & Percentage of & Method & \multicolumn{4}{c}{Training Detail} \\
Dataset & Training Data Used \\
\cmidrule(lr){4-7}
& & & \# Epochs & Batch size & Initial Learning Rate & \# Image pairs per epoch \\
\midrule
CIFAR-10 & 100\% & MSP / ODIN / & 200 & 128 & 0.1 & N/A \\
\& & & Mahalanobis / POD \\
CIFAR-100 \\
& & POD + Fine-tune & (Classifier) 175 & 128 & 0.1 & N/A \\
& & & (FT) 25 & 32 & 0.01 & 25,000 \\
\\
 & 10\% & MSP / ODIN / & 1000 & 128 & 0.1 & N/A \\
 & & Mahalanobis / POD \\
 \\
& & POD + Fine-tune & (Classifier) 750 & 128 & 0.1 & N/A \\
& & & (FT) 250 & 32 & 0.01 & 2,500 \\
\\
SVHN & 100\% & MSP / ODIN / & 200 & 128 & 0.1 & N/A \\
 & & Mahalanobis / POD \\
\\
& & POD + Fine-tune & (Classifier) 175 & 128 & 0.1 & N/A \\
& & & (FT) 25 & 32 & 0.01 & 36,000 \\
\\
SVHN & 10\% & MSP / ODIN / & 1000 & 128 & 0.1 & N/A \\
 & & Mahalanobis / POD \\
\\
& & POD + Fine-tune & (Classifier) 750 & 128 & 0.1 & N/A \\
& & & (FT) 250 & 32 & 0.01 & 3,600 \\
\bottomrule

  \end{tabular}
  \end{adjustbox}
  \caption{Training details for the neural networks we use. Note that the same neural network is used for MSP, ODIN, Mahalanobis and POD methods. For POD + Fine-tune, we first train a classifier for a fewer number of epochs, and then fine-tune it for some additional number of epochs on the pairwise classification task. For fine-tuning, a network sees $N$ same class pairs, and $N$ different class pairs - a total 2N pairs of images. We choose the number of image pairs the network sees during each epoch such that the total number of images it sees does not exceed the number of images a classifier sees during its training, to make sure POD + Fine-tune does not get any "extra" training benefit over the other methods.}
  \label{table:trainingDetails}
\end{table*}

We use a ResNet \cite{he_16} architecture with 34 layers for all our experiments (unless explicitly mentioned otherwise), similar to the works of~\citet{lee_18} and~\citet{hsu_20}. All neural networks are trained using SGD with Nesterov momentum, and with a cosine learning rate scheduler \cite{loshchilov_17}. 

We standardize the training details for each network/method, i.e., the number of epochs and the number of images a neural net sees during each epoch, so as to ensure that no method gains an unfair advantage above the other. The training detail for each method is recorded in Table \ref{table:trainingDetails}. For more details on the POD + Fine-tune method (e.g., its training algorithm), please see Appendix \ref{appendix:fineTuning}.

\subsection{Per class samples used for POD methods}

As discussed in section \ref{subsection:methods}, we need M images per each of the C classes in the In-distribution dataset for POD. Note that all of these images belong to $\mathcal{D}_{in}^{train}$ and we do not use any OOD dataset. 

For all of the experiments in this work, we choose $M = 20$. We have experimented with different values of M, and saw that using lower values of M usually harms the OOD detection performance. However, choosing $M > 20$ does not increase the performance substantially, and only increases the computational cost. 

\subsection{Low data experiment details} \label{appendix:lowDataSetup}

For our low data experiments, we use 10\% of the ID train datasets ($\mathcal{D}_{in}^{train}$) for the purpose of training and testing. For the sake of reproducibility and consistency, we form the low data train dataset ($\mathcal{D}_{in}^{train, low}$) as follows: for each of the C classes in the standard datasets, we take the first 10\% of the examples, instead of a random 10\% sample, (i.e., for CIFAR-10, we take the first 500 samples from each of the ID classes from the training dataset) and include them to form the new training dataset. We have experimented with choosing the low data subset randomly, it does not add any significant variation in the reported numbers.

\section{Discussion of OOD sample dependent hyper-parameters for ODIN / Mahalanobis} \label{appendix:additionalHyperparameterTuning}

\begin{table*}[t]
  \centering
  \begin{adjustbox}{width=1.2\columnwidth,center}
  \begin{tabular}{cccc}
     \toprule
     Method & Hyper-parameters & Search Grid & Fixed Value\\
\midrule
ODIN & Temperature (T) & 1, 10, 100, 1000 & 1000\\
& Noise ($\epsilon$) & 0, 0.0005, 0.001, 0.0014, 0.002, & 0 \\
& & 0.0024, 0.005, 0.01, \\
& & 0.05, 0.1, 0.2
\\
Mahalanobis & Noise ($\epsilon$) & 0, 0.0005, 0.001, 0.0014, 0.002, & 0 \\
& & 0.0024, 0.005, 0.01, \\
& & 0.05, 0.1, 0.2\\
\bottomrule
  \end{tabular}
  \end{adjustbox}
  \caption{Hyper-parameter search grid for ODIN \& Mahalanobis. For each (ID, OOD) pair, we evaluate the method on all pairs of $(T, \epsilon)$ from this table, and record the best results in Table \ref{table:comparisonWithOOD}. When we do not have access to OOD samples, we use the fixed values for the hyper-parameters, and those results are recorded in Table \ref{table:comparisonWithoutOOD}.}
  \label{table:hyperparameterSearchGrid}
\end{table*}

\subsection{Input pre-processing for ODIN} \label{appendix:odinParameters}

In addition to temperature scaling the softmax score by some $T > 0$ (Equation \ref{equation:S(x;T)}), ODIN \cite{liang_18} uses input pre-processing by some noise $\epsilon > 0$ to improve OOD detection. Consider input $x \in \mathcal{X}$. We define $S_{ODIN,w/o}(x)$ to the ODIN score for input $x$ without any pre-processing:

\begin{table*}[t]
  \centering
  \begin{tabular}{ccccccc}
     \toprule
     & \multicolumn{6}{c}{ID Dataset} \\
 & \multicolumn{2}{c}{CIFAR-10} & \multicolumn{2}{c}{CIFAR-100} & \multicolumn{2}{c}{SVHN} \\
\cmidrule(){2-7}
OOD Dataset & Full & Low & Full & Low & Full & Low \\
\midrule
CIFAR-10 & --- & --- & (1000, 0) & (10, 0) & (100, 0.0005) & (10, 0.005) \\
CIFAR-100 & (10, 0) & (10, 0.0005) & --- & --- & (10, 0.0005) & (10, 0.005) \\
SVHN & (10, 0) & (100, 0.05) & (10, 0.01) & (10, 0.1) & --- & --- \\
STL-10 & --- & --- & (1000, 0) & (1000, 0) & (10, 0.0005) & (10, 0.005) \\
CelebA & (1, 0.005) & (1, 0.005) & --- & --- & (10, 0.002) & (10, 0.01) \\
TinyImageNet & (1, 0.001) & (10, 0.01) & (10, 0.005) & (100, 0.01) & (10, 0.0005) & (10, 0.005) \\
LSUN & (1, 0.001) & (10, 0.01) & (10, 0.005) & (10, 0.01) & (1, 0.0014) & (10, 0.005) \\
\bottomrule
  \end{tabular}
  \caption{Optimal hyper-parameter values, $(T_{opt}, \epsilon_{opt})$, for ODIN method. Here for each (ID, OOD) pair, we evaluate ODIN with all pairs of $(T, \epsilon)$ from Table \ref{table:hyperparameterSearchGrid}, and record the pair for which ODIN obtains the highest OOD detection AUROC on that particular (ID, OOD) pair. Notice that optimal hyper-parameters change depending on whether 100\% (Full data setting) or 10\% (Low data setting) ID training data is available. This table records the optimal values for the final run of our experiments, and we note that a different run might result into different optimal hyper-parameters.}
  \label{table:odinOptimalHyperparameters}
\end{table*}

\begin{equation} \label{equation:odinWithoutPreprocessing}
    S_{ODIN,w/o}(x; T) = \max_{i \in \{1, \cdots, C\}} S_{soft}^i(x; T)
\end{equation}

where $S_{soft}^i(x; T)$ is defined in Equation (\ref{equation:S(x;T)}). Let $\hat{x} = G_{ODIN}(x; \epsilon, T)$ to be the pre-processed version of $x$. We have,

\begin{equation} \label{equation:odinPreprocessing}
    G_{ODIN}(x; \epsilon, T) = x - \epsilon \sign(-\nabla_x \log S_{ODIN,w/o}(x; T))
\end{equation}

\subsection{Class means, covariance matrix and input pre-processing for Mahalanobis} \label{appendix:mahalanobisParameters}

Note that in the original experiments from~\citet{lee_18}, one uses the weighted sum of the Mahalanobis distances from all residual blocks of a ResNet~\cite{he_16}. In our own work, we follow more recent literature (e.g., \citet{liu_20}) and use only the Mahalanobis distance of the neural network's penultimate (second-to-last) residual block's representation. The reason for this is simple ---~\citet{lee_18} assumes access to some OOD examples in order to tune the weights of individual layers. They mention that the layers that do not contribute to OOD detection have their respective weights set close to zero, and in practice become ``ineffective" layers. However, in our setting, $\mathcal{D}_{out}$ is unseen, and no OOD examples are known for training or hyper-parameter tuning, so the weights cannot be properly chosen. Hence we use only the penultimate layer.

Let $h(x)$, $\mu_i$ and $\Sigma$ be the neural network representation, class mean for label $i$, and covariance matrix at the neural network’s penultimate layer, respectively. $\mu_i$ and $\Sigma$ are defined, using the training examples $\{(x_j, y_j)\}_{j = 1}^N$ sampled from $\mathcal{D}_{in}$, as follows:

\begin{equation} \label{equation:mu}
    \mu_i = \frac{1}{N_i} \sum_{j: y_j = i} h(x_j)
\end{equation}

\begin{equation}
    \Sigma = \frac{1}{N} \sum_{i = 1}^C \sum_{j: y_j = i} (h(x_j) - \mu_i)(h(x_j) - \mu_i)^T
\end{equation}

where $N_i$ and $N$ are the number of training examples in class $i$ and the total number of training examples, respectively.

Similar to ODIN, Mahalanobis~\cite{lee_18} also employs input pre-processing by noise $\epsilon > 0$. Consider some $x \in \mathcal{X}$. We first find the closest class $c$ to example $x$ in the penultimate layer, i.e.,

\begin{equation} \label{equation:closestClass}
    c = \argmin_{i \in \{1, \cdots, C\}}  \{(h(x) - \mu_i)^T \Sigma^{-1} (h(x) - \mu_i)\}
\end{equation}

Let $\hat{x} = G_{Maha}(x ; \epsilon)$ be the pre-processed version of input $x$. We define it as follows:

\begin{equation} \label{equation:mahalanobisPreprocessing}
    \begin{split}
    G_{Maha}(x; \epsilon) = x - \epsilon \sign (\nabla_x (h(x) - \mu_i)^T \Sigma^{-1} \\ 
            (h(\hat{x}) - \mu_i))
    \end{split}
\end{equation}

\begin{figure}[!ht]
    \centering
    \includegraphics[width=\columnwidth]{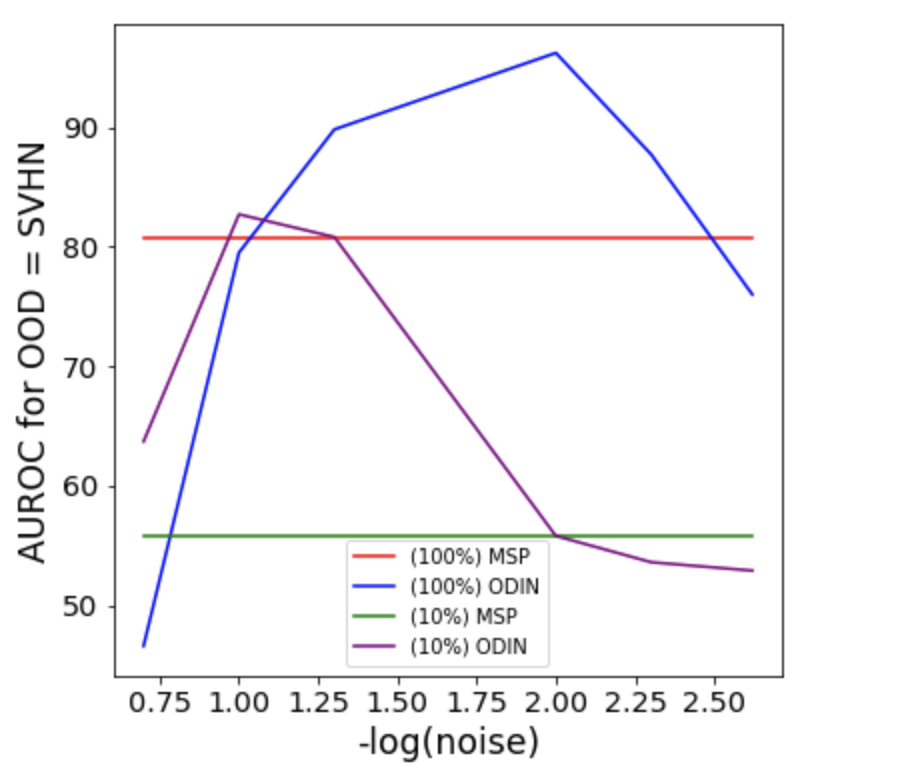}
    \caption{ODIN's sensitivity to noise magnitude, when ID = CIFAR-100, OOD = SVHN, and temperature $T = 1000$. Here we plot the OOD detection AUROC vs $-\log_{10}(\epsilon)$. The horizontal red and green lines are MSP baselines for when 100\% and 10\% ID train data is available, respectively. Notice that the best performing noise for the full data regime results in very poor performance in the low data regime and vice versa.}
    \label{fig:odinSensitivity}
\end{figure}

\subsection{OOD sample dependent hyper-parameters for ODIN}

The hyper-parameters $T$ and $\epsilon$ for ODIN require OOD samples to be tuned. In Table \ref{table:comparisonWithOOD}, we evaluate ODIN on each (ID, OOD) pair using a pre-defined search grid for $T$ and $\epsilon$, and record the best performance (in terms of AUROC). We use the same architecture (ResNet-34) as~\citet{lee_18}, and so we also use their search grid, as recorded in Table \ref{table:hyperparameterSearchGrid}.

However, we also consider the case when $\mathcal{D}_{out}$ is completely unseen. In that case, we used the fixed values for $(T, \epsilon)$ as recorded in Table \ref{table:hyperparameterSearchGrid}. 

It is mentioned in the original works by~\citet{lee_18} that higher values of $T$ are to be preferred, but too high values do not yield any additional benefits. Following their work and~\citet{hsu_20}, we set $T = 1000$ when we do not have access to OOD samples. 

However, unlike temperature, there is no principled way for setting $\epsilon$ without access to OOD samples. We see that the performance of ODIN, specially on the low data regime, is highly sensitive to the choice of $\epsilon$, i.e., slight changes in $\epsilon$ can harm the performance, and the optimal $\epsilon$ depends on too many factors. Even when we keep $\mathcal{D}_{in}$ fixed, the optimal choice of $\epsilon$ varies for different choices of $\mathcal{D}_{out}$ and the individual ID training samples available. We demonstrate this in Figure \ref{fig:odinSensitivity}, where for (ID, OOD) = (CIFAR-100, SVHN) and $T = 1000$, we see that using the best performing noise for the full data regime results in poor performance in the low data regime and vice versa.

So the optimal $\epsilon$ for input pre-processing needs to be tuned for each neural network architecture~\cite{lee_18}, individual (ID, OOD) pair and the examples sampled. Setting $\epsilon$ to some arbitrary value above 0 without tuning it properly can make ODIN under-perform MSP. This is why in the setting where we do not have access to OOD samples, we simply set $\epsilon = 0$.

\subsection{OOD sample dependent hyper-parameters for Mahalanobis}

The hyper-parameter $\epsilon$ for Mahalanobis, similar to ODIN, requires OOD samples to be tuned. We use the same architecture (ResNet-34) as~\citet{lee_18}, and so we also use their search grid, as recorded in Table \ref{table:hyperparameterSearchGrid}.

Similar to ODIN, there is no principled way of choosing $\epsilon$, and we set $\epsilon = 0$ in the setting without any OOD samples.

\begin{table}[t]
  \centering
  \begin{adjustbox}{width=\columnwidth,center}
  \begin{tabular}{ccccccc}
     \toprule
     & \multicolumn{6}{c}{ID Dataset} \\
 & \multicolumn{2}{c}{CIFAR-10} & \multicolumn{2}{c}{CIFAR-100} & \multicolumn{2}{c}{SVHN} \\
\cmidrule(){2-7}
OOD Dataset & Full & Low & Full & Low & Full & Low \\
\midrule
CIFAR-10 & --- & --- & 0 & 0.01 & 0.005 & 0.01 \\
CIFAR-100 & 0 & 0.01 & --- & --- & 0.005 & 0.005 \\
SVHN & 0.01 & 0.01 & 0.01 & 0.05 & --- & --- \\
STL-10 & --- & --- & 0 & 0.01 & 0.01 & 0.01 \\
CelebA & 0 & 0.01 & --- & --- & 0.01 & 0.01 \\
TinyImageNet & 0.0024 & 0.005 & 0.005 & 0.01 & 0.01 & 0.01 \\
LSUN & 0.0024 & 0.005 & 0.0024 & 0.01 & 0.01 & 0.01\\
\bottomrule
  \end{tabular}
  \end{adjustbox}
  \caption{Optimal hyper-parameter values, $\epsilon_{opt}$, for Mahalanobis method. Here for each (ID, OOD) pair, we evaluate Mahalanobis with all values of $\epsilon$ from Table \ref{table:hyperparameterSearchGrid}, and record the value for which Mahalanobis obtains the highest OOD detection AUROC on that particular (ID, OOD) pair.}
  \label{table:mahalanobisOptimalHyperparameters}
\end{table}

\subsection{Optimal hyper-parameters for ODIN}

For each (ID, OOD) pair, we evaluate ODIN with all pairs of $(T, \epsilon)$ from Table \ref{table:hyperparameterSearchGrid}, and record the pair for which ODIN obtains the highest OOD detection AUROC on that particular (ID, OOD) pair, $(T_{opt}, \epsilon_{opt})$, in Table \ref{table:odinOptimalHyperparameters}. 

Note that these are the optimal hyper-parameters of our final run based on which we produce the results in this paper, and there can be variations in these numbers in different runs.

\subsection{Optimal hyper-parameters for Mahalanobis}

Similar to Table \ref{table:odinOptimalHyperparameters}, we record the optimal hyper-paramters for Mahalanobis, $\epsilon_{opt}$, for each (ID, OOD) pair in Table \ref{table:mahalanobisOptimalHyperparameters}.

\section{More on the toy datasets} \label{appendix:toyDatasets}

\subsection{Algorithms for generating the toy datasets} \label{appendix:algorithmsForToyDatasets}

Here we discuss the pseudocode of a few functions that we later use to generate our toy datasets. Specifically, first we write the pseudocode to create a cluster of data-points in $\mathbb{R}^2$ around $(\mu_x, \mu_y)$ in Algorithm \ref{algorithm:cluster}.

\begin{algorithm}
   \caption{produceCluster}
   \label{algorithm:cluster}
\begin{algorithmic}
   \STATE {\bfseries Input:} Mean $(\mu_x, \mu_y)$, standard deviations $(\sigma_x, \sigma_y)$, number of data-points $N$
   \STATE {\bfseries Initialize} $L$ to be an empty list
   \FOR{$i = 1$ {\bfseries to} $N$}
       \STATE Pick $x \sim \mathcal{N}(\mu_x, \sigma_x^2)$ and $y \sim \mathcal{N}(\mu_y, \sigma_y^2)$
       \STATE Append $(x, y)$ to the list $L$
   \ENDFOR
   \STATE {\bfseries Return} $L$
\end{algorithmic}
\end{algorithm}

Next we write pseudocode to create a dataset in $\mathbb{R}^2$, where the data-points are spread around a circle. The pseudocode is presented in Algorithm \ref{algorithm:ring}.

\begin{algorithm}
   \caption{produceRing}
   \label{algorithm:ring}
\begin{algorithmic}
   \STATE {\bfseries Input:} Center $(C_x, C_y)$, radius $R$, number of points $N$
   \STATE {\bfseries Initialize} $L$ to be an empty list
   \FOR{$k = 1$ {\bfseries to} $N$}
       \STATE Let $\theta = \frac{2\pi k}{N}$
       \STATE Let $x = C_x + R\cos{\theta}$, $y = C_y + R\sin{\theta}$
       \STATE Append $(x, y)$ to the list $L$
   \ENDFOR
   \STATE {\bfseries Return} $L$
\end{algorithmic}
\end{algorithm}

\subsection{Generating toy dataset 1} \label{appendix:generatingToyDataset_1}

We use Algorithm $\ref{algorithm:generateToyDataset_1}$ to produce toy dataset 1 in Figure \ref{fig:toyDataset1}. Note that we randomly choose 80\% of the ID examples to form dataset $\mathcal{D}_{in}^{train}$, and the other 20\% to form dataset $\mathcal{D}_{in}^{test}$.

\begin{algorithm}
   \caption{generateToyDataset1}
   \label{algorithm:generateToyDataset_1}
\begin{algorithmic}
   \STATE Let ID\_Class\_1 = produceCluster($\mu_x = -10$, $\mu_y = -10$, $\sigma_x = 0.5$, $\sigma_y = 0.5$, N = 100)
   \STATE Let ID\_Class\_2 = produceCluster($\mu_x = -2$, $\mu_y = -2$, $\sigma_x = 0.5$, $\sigma_y = 0.5$, N = 100)
   \STATE Let OOD\_Class = produceCluster($\mu_x = 6.5$, $\mu_y = 6.5$, $\sigma_x = 0.5$, $\sigma_y = 0.5$, N = 40)
   \STATE {\bfseries Return} ID\_Class\_1, ID\_Class\_2, OOD\_Class
\end{algorithmic}
\end{algorithm}

\subsection{Generating toy dataset 2} \label{appendix:generatingToyDataset_2}

We use Algorithm $\ref{algorithm:generateToyDataset_2}$ to produce toy dataset 2 in Figure \ref{fig:toyDataset2}. We randomly choose 50\% of the ID examples to form dataset $\mathcal{D}_{in}^{train}$, and the other 50\% to form dataset $\mathcal{D}_{in}^{test}$.

\begin{algorithm}
   \caption{generateToyDataset2}
   \label{algorithm:generateToyDataset_2}
\begin{algorithmic}
   \STATE Let ring\_1 = produceRing($C_x = -5$, $C_y = -5$, $R = 2.5$, $N = 9$
   \STATE Let cluster\_1 = produceCluster($\mu_x = -0.8$, $\mu_y = -0.8$, $\sigma_x = 0.2$, $\sigma_y = 0.2$, N = 6)
   \STATE Let ID\_Class\_1 = ring\_1 $\cup$ cluster\_1 \\
   \STATE Let ID\_Class\_2 = produceRing($C_x = 8$, $C_y = 8$, $R = 2.5$, $N = 15$)
   \STATE Let OOD\_Class = produceCluster($\mu_x = 0$, $\mu_y = 0$, $\sigma_x = 0.3$, $\sigma_y = 0.3$, N = 10)
   \STATE {\bfseries Return} ID\_Class\_1, ID\_Class\_2, OOD\_Class
\end{algorithmic}
\end{algorithm}

\section{More on Pairwise OOD Detection (POD)} \label{appendix:pod}

\subsection{Scoring mechanism} \label{appendix:accumulators}

Here we consider a more general setting for the Pairwise OOD Detection (POD) method than that described in Section \ref{subsection:methods}, and describe why we choose the special setting in Section \ref{subsection:methods} even though there were other options available. 

Consider the problem description in Section \ref{section:problemDescription}. Let $h(x) \in \mathbb{R}^d$ be the neural network's penultimate layer’s representation for test sample $x$. Suppose each of the $C$ ID classes has $M$ training examples i.e., $\{\{z_{ij}\}_{j=1}^M\}_{i=1}^C$ where $z_{ij}$ is the $j$-th example from the $i$-th ID class (Here we consider all ID train examples, however in practice some smaller value of M is chosen to reduce computation costs). Let $g_{cls}$ and $g_{ex}$ be two functions from an arbitrary sized non-empty set of real numbers to $\mathbb{R}$ --- for example, $g_{cls}$ or $g_{ex}$ can be the max, min or average function. We use $g_{ex}$ to accumulate the distance from a test example to all the training examples from a particular class, and $g_{cls}$ to accumulate all such class-specific scores to obtain the final score for the test example. Specifically, for any $x \in \mathcal{X}$, we define the following:
\begin{equation} \label{equation:pairwiseDistance}
    dist(x, p_{ij}) = ||h(x) - h(z_{ij})||_2^2
\end{equation}

\begin{equation} \label{equation:generalPODDistance}
    S^i_{POD}(x) = g_{ex}(\{dist(x, z_{ij})\}_{j = 1}^M)
\end{equation}

Here instead of taking the average of $dist(x, p_{ij})$ over $j = 1$ to $M$, we use the general function $g_{ex}$. Similarly, we define,

\begin{equation} \label{equation:generalPODScore}
    S_{POD}(x) = g_{cls}(\{S^i_{POD}(x)\}_{i = 1}^C)
\end{equation}

The choice for the function pair $(g_{cls}, g_{ex})$ constitutes the scoring scheme for the POD method. We note that there are a couple intuitive choices for $(g_{cls}, g_{ex})$ such as (min, average), (average, average) and (min, min). In all our experiments, we use (min, average). Here we provide a toy dataset first showing why we make this choice, next we provide empirical evidence supporting our choice.

\subsection{Toy dataset differentiating between different POD scoring schemes}

Consider toy dataset 2 in Figure \ref{fig:toyDataset2}. In the original example, we consider $(g_{cls}, g_{ex}) = $ (min, min) and show that distance based methods perform poorly on the dataset. Here however, we consider different intuitive choices for $(g_{cls}, g_{ex})$ and record their performances in Table \ref{table:accumulatorToyDataset}.

\begin{threeparttable}
  \centering
  \begin{adjustbox}{width=0.7\columnwidth,center}
  \begin{tabular}{cc}
     \toprule
     Choice for $(g_{cls}, g_{ex})$ & OOD detection AUROC \\
\midrule
(min, min) & 32.5\% \\
(min, average) & 100.0\% \\
(average, average) & 0\% \\
\bottomrule

  \end{tabular}
  \end{adjustbox}
  \caption{Performance of POD method (OOD Detection AUROC) on toy dataset 2 under different choices for $(g_{cls}, g_{ex})$.}
  \label{table:accumulatorToyDataset}
\end{threeparttable}

We have already discussed that the ID outliers force (min, min) to perform poorly. However, (min, average) does not have that flaw, under the assumption that outliers constitute only a small portion of the ID dataset. This is because averaging over the training examples in individual classes negates the effect of the few ID outliers. (average, average) does the worst here because the OOD class is actually between the two ID classes. For a test example belonging to ID class 1, the distances to ID class 2 examples will be much larger and vice versa, making the average distance large for ID test examples --- however, the OOD samples are between both classes, and so the average distance is smaller for OOD samples.

\subsection{Empirical evidence for the choice of POD scoring scheme}

We also observe empirically in our experiments that (min, average) performs most consistently across different (ID, OOD) pairs and full vs low data experiments. We record the OOD detection performance of POD method for different choices of $(g_{cls}, g_{ex})$ in Table \ref{table:accumulatorEmpiricalResults} when ID = CIFAR-10 and 10\% training data is used.

\begin{threeparttable}
  \centering
  \begin{adjustbox}{width=\columnwidth,center}
  \begin{tabular}{cccc}
     \toprule
     & \multicolumn{3}{c}{OOD detection AUROC for choice of $(g_{cls}, g_{ex})$} \\
\cmidrule(){2-4} 
OOD Dataset & (min, average) & (average, average) & (min, min) \\
\midrule
CIFAR-100 & \textbf{77.9} & 77.4 & 77.1 \\
SVHN & \textbf{76.2} & 44.1 & 71.0 \\
CelebA & \textbf{77.7} & 75.9 & 74.4 \\
TinyImageNet & 73.0 & \textbf{75.9} & 76.2 \\
LSUN & 77.4 & \textbf{82.7} & 80.7 \\
\bottomrule
  \end{tabular}
  \end{adjustbox}
  \caption{OOD detection performance of POD method with ID = CIFAR-10 and low data setting (10\% ID train data used) for different choices of $(g_{cls}, g_{ex})$. Bold numbers represent superior result on a particular (ID, OOD) pair.}
  \label{table:accumulatorEmpiricalResults}
\end{threeparttable}

However, we note that different $(g_{cls}, g_{ex})$ choices work better on different (ID, OOD) pairs, and the inconsistency among these results are also significant.. This can be considered a hyper-parameter for POD method, and we suggest investigating more into how conceptually/empirically the choice of $(g_{cls}, g_{ex})$ affects OOD detection. In the more structured setting where one has access to small number of OOD examples, one may also follow ODIN~\cite{liang_18} or Mahalanobis~\cite{lee_18} and use OOD samples to determine the choice for $(g_{cls}, g_{ex})$. 

\section{Fine-tuning on pairwise classification task}

Inspired by the OOD detection performance of the POD method, we also experiment with fine-tuning a regular classifier on the pairwise classification task. We call this method POD + Fine-tune. We will first discuss the algorithm for fine-tuning, and then discuss the findings. 

\subsection{Algorithm for fine-tuning} \label{appendix:fineTuning}

Let us assume that the penultimate layer of the regular neural network classifier, $h$, produces an embedding in $\mathbb{R}^d$, i.e., for any $x \in \mathcal{X}$, we have $h(x) \in \mathbb{R}^d$. For fine-tuning, we introduce a learnable parameter $w \in \mathbb{R}^d$. Given an image pair $(x_1, x_2) \in \mathcal{X} \times \mathcal{X}$, and label $y_p = 0$ if $x_1$ and $x_2$ belong to the same class, or $y_p = 1$, we define the model output as follows:

\begin{equation}
    Pairwise(x, y) = \sum_{j = 1}^d w_j(h(x_1)_j - h(x_2)_j)^2
\end{equation}

We fine-tune $f$ and $w$ with binary cross entropy loss between $y_p$ and $Pairwise(x, y)$ using Algorithm \ref{algorithm:fineTune}.

\newpage

\begin{algorithm}
  \caption{Fine Tune POD}
  \label{algorithm:fineTune}
\begin{algorithmic}
  \STATE {\bfseries Input:} Regular classifier $f$, Train dataset $\mathcal{D}_{train}$, number of epochs $N_{epoch}$, number of data-pairs per epoch $N_{pairs}$
  \FOR{$i = 1$ {\bfseries to} $N_{epoch}$}
      \STATE {\bfseries Initialize} $\mathcal{D}_p$ to be an empty dataset
      \STATE {\bfseries Initialize} $w$ to be a vector in $\mathbb{R}^d$
      \FOR{$j = 1$ {\bfseries to} $N_{pairs}$}
            \IF{$j$ is even}
                \STATE Randomly choose a class $c$ in $\mathcal{D}_{train}$
                \STATE Randomly choose data-points $(x_1, y_1)$, $(x_2, y_2) \in \mathcal{D}_{train}$ such that $x_1 \neq x_2$ and $y_1 = y_2 = c$
                \STATE Append $((x_1, x_2), 0)$ to $\mathcal{D}_p$
            
            \ELSE
                \STATE Randomly choose classes $c_1 \neq c_2$ in $\mathcal{D}_{train}$
                \STATE Randomly choose data-point $(x_1, y_1) \in \mathcal{D}_{train}$ such that $y_1 = c_1$
                \STATE Randomly choose data-point $(x_2, y_2) \in \mathcal{D}_{train}$ such that $y_2 = c_2$
                \STATE Append $((x_1, x_2), 1)$ to $\mathcal{D}_p$
            \ENDIF
      \ENDFOR
      \STATE {\bfseries Fine-tune} $f$ and $w$ on dataset $\mathcal{D}_p$
  \ENDFOR
\end{algorithmic}
\end{algorithm}

Following the setting described for POD method in Section \ref{subsection:methods}, we define:

\begin{equation} \label{equation:FT_class_i}
    S_{POD+FT}^i(x) = \frac{1}{M} \sum_{j = 1}^M \sum_{k = 1}^d w_k(h(x)_k - h (z_{ij})_k)^2
\end{equation}

Finally, we saw in Appendix \ref{appendix:pod} that (min, average) is the best scoring scheme for POD. Surprisingly, (average, average) is the overall best scoring scheme for POD + Fine-tune (See empirical results for ID = CIFAR-10 and the full data setting on Table \ref{table:accumulatorPOD_FT}). We suggest further investigations into why this occurs. With this in mind, we define the score function for POD + Fine-tune as follows:
\begin{equation} \label{equation:POD_FT_Score}
    S_{POD+FT}(x) = \frac{1}{C} \sum_{i = 1}^C S_{POD,FT}^i(x)
\end{equation}

And use this as a heuristic for OOD detection as described in Section \ref{section:problemDescription}.

\subsection{Observation}
We notice that POD + Fine-tune actually does worse than POD in most of the cases (See Table \ref{table:fullDataResults} and \ref{table:lowDataResults}). Even though this is surprising, we note that it is similar to the findings of~\citet{reed_21} and ~\citet{kolesnikov_19}.~\citet{kolesnikov_19} finds that
directly training a full network on rotation prediction harms the correlation between rotation prediction and supervised performance. Additionally,~\citet{reed_21} shows that training a 2-layer network on rotation prediction improves the performance on the rotation prediction task but hurts the correlation between rotation accuracy and supervised evaluation of learned representations. Similarly, in our work we see that fine-tuning a whole ResNet-34~\cite{he_16} network on pairwise prediction task improves pairwise accuracy --- but since the network can learn its own representations, this ends up hurting OOD detection performance in most cases. This may require more study in the future.

\begin{table}
  \centering
  \begin{adjustbox}{width=\columnwidth,center}
  \begin{tabular}{cccc}
     \toprule
     & \multicolumn{3}{c}{OOD detection AUROC for choice of $(g_{cls}, g_{ex})$} \\
\cmidrule(){2-4} 
OOD Dataset & (min, average) & (average, average) & (min, min) \\
\midrule
CIFAR-100 & 88.3 & 87.1 & \textbf{88.6} \\
SVHN & 93.8 & \textbf{97.0} & 92.3 \\
CelebA & 73.0 & \textbf{77.6} & 72.1 \\
TinyImageNet & 91.2 & \textbf{92.6} & 92.3 \\
LSUN & 93.3 & \textbf{94.5} & 94.0 \\
\bottomrule
  \end{tabular}
  \end{adjustbox}
  \caption{OOD detection performance of POD + Fine-tune method with ID = CIFAR-10 and full data setting (100\% ID train data used) for different choices of $(g_{cls}, g_{ex})$ (See Appendix \ref{appendix:accumulators} for details of scoring mechanism). Bold numbers represent superior result on a particular (ID, OOD) pair.}
  \label{table:accumulatorPOD_FT}
\end{table}

\section{ID classification accuracy} \label{appendix:idClassificationAccuracy}

We also record the in-distribution classification accuracy of MSP, POD and POD + Fine-tune methods. Given $x \in \mathcal{X}$, we define ---

\begin{equation}
    Pred_{MSP}(x) = \argmax_{i \in \{1, \cdots, C\}} S^i_{soft}(x; T = 1)
\end{equation}

\begin{equation}
    Pred_{POD}(x) = \argmin_{i \in \{1, \cdots, C\}} S^i_{POD}(x)
\end{equation}

\begin{equation}
    Pred_{POD+FT}(x) = \argmin_{i \in \{1, \cdots, C\}} S^i_{POD+FT}(x)
\end{equation}

where $S^i_{soft}(x; T)$, $S^i_{POD}(x)$ and $S^i_{POD}(x)$ are defined in Equations \ref{equation:S(x;T)}, \ref{equation:pairwise_class_i} and \ref{equation:FT_class_i}, respectively. The results are summarized in Table \ref{table:idClassificationAccuracy}.

\begin{threeparttable}
  \centering
  \begin{adjustbox}{width=\columnwidth,center}
  \begin{tabular}{ccccccc}
     \toprule
     & \multicolumn{2}{c}{MSP} & \multicolumn{2}{c}{POD} & \multicolumn{2}{c}{POD + FT} \\
\cmidrule{2-7}
ID Dataset & Full & Low & Full & Low & Full & Low \\
\midrule
CIFAR-10 & 95.75 & 82.21 & 95.77 & 81.74 & 94.42 & 83.03 \\
CIFAR-100 & 79.26 & 37.37 & 79.17 & 36.63 & 72.56 & 37.55 \\
SVHN & 96.12 & 91.95 & 96.09 & 91.93 & 95.98 & 91.26 \\
\bottomrule
  \end{tabular}
  \end{adjustbox}
  \caption{Classification accuracy on $\mathcal{D}_{in}^{test}$, for MSP, POD and POD + FT methods.}
  \label{table:idClassificationAccuracy}
\end{threeparttable}

\begin{figure*}[ht]
    \centering
    \begin{minipage}[b]{0.40\linewidth}
         \centering
         \includegraphics[width=\columnwidth]{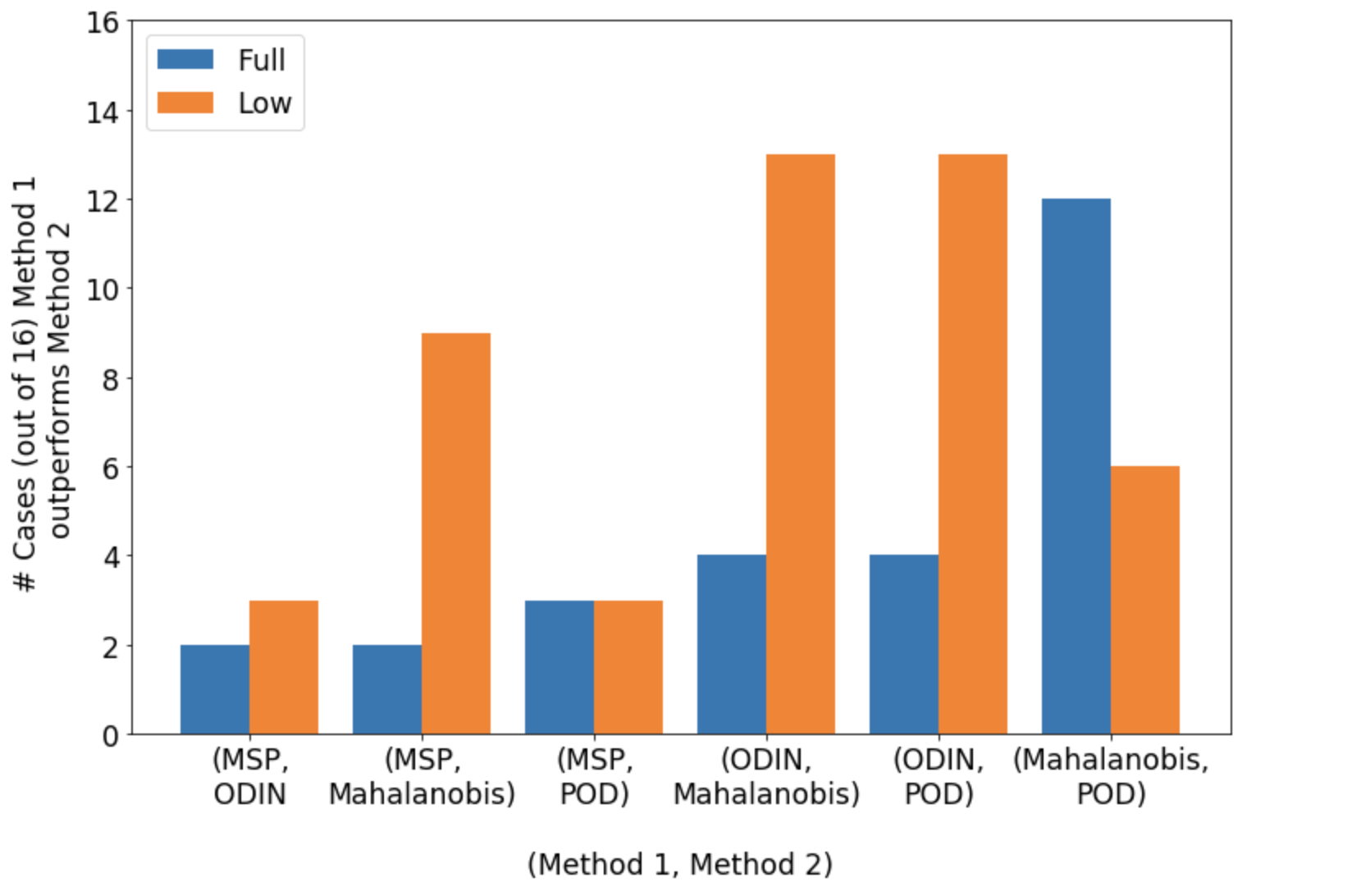}
         
         \caption{Bar plot showing the number of ID-OOD dataset pairs (out of 16) on which Method 1 outperforms Method 2 on OOD detection by AUROC, without any access to OOD samples for training/hyper-parameter tuning. Notice the change in performance due to the change in ID training dataset size. In 4 out of 6 cases, one method outperforming another on the majority of (ID, OOD) pairs in the full data setting does not do so in the low data regime.}
         \label{fig:comparisonWithoutOODSamples}
    \end{minipage}
    \quad
    \begin{minipage}[b]{0.40\linewidth}
    \centering
         \includegraphics[width=\columnwidth]{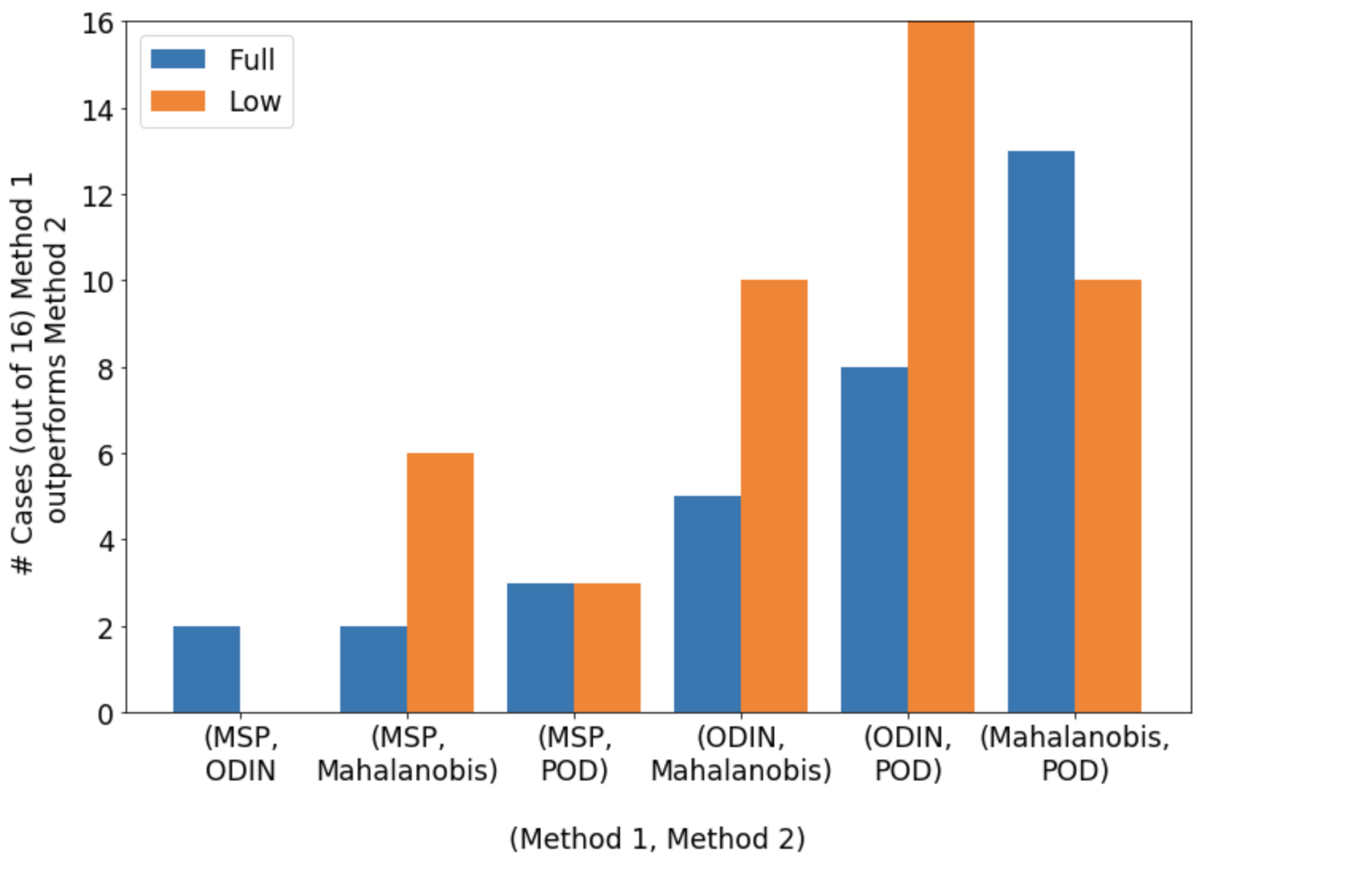}
         
         \caption{Bar plot showing the number of ID-OOD dataset pairs (out of 16) on which Method 1 outperforms Method 2 on OOD detection by AUROC, where ODIN and Mahalanobis have access to OOD samples for training/hyper-parameter tuning. Despite inconsistency being lower compared to Figure \ref{fig:comparisonWithoutOODSamples}. However, there is still inconsistency, for example the performance order on the majority of 16 cases between ODIN and Mahalanobis switches from low to full data setting.} 
         \label{fig:comparisonWithOODSamples}
    \end{minipage}
\end{figure*}

We see that POD maintains very similar classification performance as MSP in all cases, however, POD + FT has performance deterioration in some cases. This might be because the neural net prioritizes learning embeddings that improves pairwise classification performance at the cost of regular classification performance during fine-tuning. 

\section{Empirical results}

For full data setting (100\% ID train data is available), we record our detailed results in Table \ref{table:fullDataResults}. Similarly we record our low data setting (10\% ID train data is available) experiment results in Table \ref{table:lowDataResults}. In both cases, we record the (AUROC, FNR@95) pair for each (ID, OOD) pair and each method. We compare between two methods based on their OOD detection AUROC first, and then break any ties using FNR@95.

\begin{table*}
  \centering
  \begin{adjustbox}{width=\textwidth,center}
  \begin{tabular}{ccccccccc}
     \toprule
     & & \multicolumn{7}{c}{OOD Detection Performance --- (AUROC, FNR@95)} \\
\cmidrule{3-9}
ID & OOD & MSP & ODIN (W/O OOD) & ODIN (With OOD) & Maha (W/O OOD) & Maha (With OOD) & POD & POD + FT \\
\midrule
CIFAR-10 & CIFAR-100 & (86.7, 68.9) & (85.6, 77.7) & (85.8, 76.3) & \textbf{(89.7, 39.8)} & \textbf{(89.7, 39.8)} & (89.3, 40.9) & (87.1, 62.1) \\

& SVHN & (95.0, 13.7) & (95.7, 12.9) & (95.7, 12.9) & (95.1, 11.9) & \textbf{(98.2, 9.4)} & (94.7, 15.0) & (97.0, 13.6) \\

& CelebA & (74.3, 78.4) & (67.6, 87.0) & (72.5, 79.6) & (76.0, 69.6) & (76.0, 69.6) & (77.0, 63.1) & \textbf{(77.6, 58.3)} \\

& TinyImageNet & (90.0, 53.8) & (90.4, 63.1) & (90.8, 62.3) & (93.1, 23.0) & \textbf{(95.3, 21.0)} & (92.3, 29.3) & (92.6, 38.0) \\

& LSUN & (93.2, 21.3) & (94.1, 23.7) & (94.5, 26.3) & (94.9, 15.6) & \textbf{(97.1, 11.5)} & (94.1, 20.3) & (94.5, 25.4) \\

\midrule
CIFAR-100 & CIFAR-10 & (78.1, 64.5) & \textbf{(78.5, 66.6)} & \textbf{(78.5, 66.6)} & (75.3, 71.2) & (75.3, 71.2) & (76.0, 76.9) & (76.2, 69.6) \\

& SVHN & (80.7, 50.5) & (81.2, 51.1) & \textbf{(96.2, 17.8)} & (88.1, 41.4) & (94.3, 23.8) & (83.7, 48.3) & (76.7, 57.6) \\

& STL-10 & (79.4, 59.4) & \textbf{(79.8, 60.8)} & \textbf{(79.8, 60.8)} & (77.6, 64.1) & (77.6, 64.1) & (78.1, 68.0) & (77.4, 65.9) \\

& TinyImageNet & (81.2, 53.3) & (83.1, 51.2) & \textbf{(90.1, 41.6)} & (83.7, 50.0) & (89.3, 41.9) & (84.3, 48.4) & (87.3, 39.2) \\

& LSUN & (82.7, 48.6) & (84.7, 46.8) & (91.5, 35.1) & (86.9, 41.8) & \textbf{(91.8, 32.4)} & (86.8, 42.1) & (90.1, 33.7) \\

\midrule
SVHN & CIFAR-10 & (95.5, 12.2) & (95.9, 12.0) & (96.0, 11.8) & (96.3, 11.7) & \textbf{(97.8, 8.1)} & (96.2, 11.9) & (93.8, 34.5) \\

& CIFAR-100 & (95.0, 13.3) & (95.5, 13.3) & (95.5, 13.3) & (96.0, 12.4) & \textbf{(97.4, 9.1)} & (95.9, 12.7) & (93.0, 39.9) \\

& STL-10 & (95.7, 11.4) & (96.1, 11.1) & (96.2, 10.9) & (96.4, 11.1) & \textbf{(98.0, 8.4)} & (96.3, 11.3) & (94.7, 26.5) \\

& CelebA & (96.5, 10.0) & (97.0, 9.6) & (97.5, 9.8) & (97.0, 9.9) & \textbf{(98.7, 5.3)} & (96.8, 10.6) & (94.1, 29.5) \\

& TinyImageNet & (95.2, 12.5) & (95.6, 12.5) & (95.7, 12.2) & (96.1, 11.5) & \textbf{(98.1, 7.8)} & (96.1, 11.8) & (93.7, 34.1) \\

& LSUN & (94.8, 13.5) & (95.2, 13.7) & (95.3, 13.8) & (96.0, 12.3) & \textbf{(98.1, 7.7)} & (95.9, 12.6) & (93.1, 37.4) \\
\bottomrule

  \end{tabular}
  \end{adjustbox}
  \caption{Detailed results for full data setting (100\% ID training data available). For each (ID, OOD) pair and for each method, we report the (AUROC, FNR@95) pair. Bold results represent the superior result for a particular (ID, OOD) pair. We compare between two methods based on their OOD detection AUROC first, and then break any ties using FNR@95. All numbers are average of 3 runs.}
  \label{table:fullDataResults}
\end{table*}

\begin{table*}
  \centering
  \begin{adjustbox}{width=\textwidth,center}
  \begin{tabular}{ccccccccc}
     \toprule
     & & \multicolumn{7}{c}{OOD Detection Performance --- (AUROC, FNR@95)} \\
\cmidrule{3-9}
ID & OOD & MSP & ODIN (W/O OOD) & ODIN (With OOD) & Maha (W/O OOD) & Maha (With OOD) & POD & POD + FT \\
\midrule
CIFAR-10 & CIFAR-100 & (77.8, 59.3) & (79.0, 61.6) & (79.1, 61.7) & (72.7, 70.7) & (73.5, 72.6) & (77.9, 63.4) & \textbf{(80.2, 59.2)} \\

& SVHN & (60.5, 84.0) & (55.5, 86.2) & \textbf{(93.5, 31.3)} & (89.7, 39.1) & (93.4, 31.6) & (76.2, 51.7) & (76.2, 59.1) \\

& CelebA & (75.4, 57.2) & (75.0, 60.4) & \textbf{(78.4, 57.5)} & (65.9, 74.7) & (68.9, 68.2) & (77.7, 55.6) & (74.6, 50.1) \\

& TinyImageNet & (74.7, 65.2) & (77.6, 63.4) & \textbf{(88.7, 43.5)} & (80.9, 57.3) & (87.9, 48.0) & (73.0, 73.0) & (74.6, 71.7) \\

& LSUN & (81.2, 47.7) & (84.8, 44.4) & \textbf{(93.9, 24.2)} & (82.0, 49.9) & (90.0, 36.6) & (77.4, 59.5) & (81.7, 55.1)\\

\midrule
CIFAR-100 & CIFAR-10 & (61.0, 82.1) & (62.0, 82.2) & \textbf{(62.1, 82.2)} & (55.3, 87.1) & (56.0, 87.1) & (61.4, 80.8) & (61.0, 82.9) \\

& SVHN & (55.8, 82.8) & (52.8, 83.4) & \textbf{(82.7, 54.4)} & (82.3, 66.0) & (82.7, 56.1) & (62.0, 82.7) & (49.8, 86.6) \\

& STL-10 & (62.0, 82.1) & \textbf{(62.5, 81.2)} & \textbf{(62.5, 81.2)} & (54.9, 87.0) & (55.2, 86.8) & (61.4, 81.2) & (61.2, 82.3) \\

& TinyImageNet & (65.0, 80.7) & (66.8, 79.2) & \textbf{(69.6, 78.1)} & (52.0, 87.0) & (59.4, 83.2) & (65.1, 78.4) & (62.7, 80.2) \\

& LSUN & (66.1, 79.3) & (68.4, 76.6) & \textbf{(70.6, 75.1)} & (50.9, 86.5) & (58.0, 83.6) & (65.5, 77.6) & (60.6, 83.1) \\

\midrule
SVHN & CIFAR-10 & (92.4, 22.0) & (93.2, 21.6) & (94.9, 18.4) & (92.4, 21.5) & \textbf{(95.3, 16.8)} & (92.4, 21.9) & (91.6, 25.9) \\

& CIFAR-100 & (92.1, 22.7) & (92.8, 22.8) & (94.2, 20.4) & (92.1, 22.5) & \textbf{(94.6, 18.8)} & (92.1, 22.4) & (91.6, 26.5) \\

& STL-10 & (92.5, 21.6) & (93.5, 21.1) & (95.2, 17.6) & (92.4, 21.8) & \textbf{(95.6, 16.3)} & (92.6, 21.4) & (91.9, 25.4) \\

& CelebA & (94.0, 17.3) & (95.0, 16.4) & (97.4, 10.0) & (93.8, 17.9) & \textbf{(97.4, 9.8)} & (94.2, 17.8) & (91.9, 26.0) \\

& TinyImageNet & (92.5, 21.4) & (93.3, 21.1) & (95.3, 17.5) & (92.5, 21.0) & \textbf{(95.7, 16.1)} & (92.5, 21.2) & (92.1, 25.0) \\

& LSUN & (92.2, 22.1) & (92.9, 21.9) & (95.3, 17.4) & (92.1, 22.0) & (95.8, 15.8) & (92.5, 21.4) & (91.8, 25.4) \\
\bottomrule

  \end{tabular}
  \end{adjustbox}
  \caption{Detailed results for low data setting (10\% ID training data available). For each (ID, OOD) pair and for each method, we report the (AUROC, FNR@95) pair. Bold results represent the superior result for a particular (ID, OOD) pair. We compare between two methods based on their OOD detection AUROC first, and then break any ties using FNR@95. All numbers are average of 3 runs.}
  \label{table:lowDataResults}
\end{table*}

\section{Full vs low data regime} \label{appendix:fullVsLowComparison}

One major inconsistency that we notice is between the full (100\%) and low (10\%) data regimes. This inconsistency is most pronounced when there is no access to OOD samples (see \ref{fig:comparisonWithoutOODSamples}), and we see that in 4 out of 6 total (method 1, method 2) pairs, if method 1 outperforms method 2 on the majority of (ID, OOD) pairs in the full data setting, than the situation reverses on low data setting. This is concerning, because in a real world setting, it would be hard to know if we have enough data to capture the diversity in the ID dataset. Access to large amount of ID data is often difficult to have --- in scenarios like monitoring bio-diversity ~\cite{hoye_21} or classifying diseases ~\cite{prabhu_18} --- due to a variety of reason, most notably, difficulty in data collection and regulations due to confidentiality.

We see that without large number of ID data, a method performing well may not do so. This problem is mitigated a little bit if there is access some OOD examples to tune hyper-parameters, as we see in Figure \ref{fig:comparisonWithOODSamples}. However, there is still inconsistency. For example, with access to OOD samples, ODIN does better than Mahalanobis on only 5 out of 16 (ID, OOD) pairs in the full data setting; however, in the low data setting, ODIN does better than Mahalanobis on 10 out of 16 pairs. This shows that we should consider another factor while discussing the OOD detection problem, namely the size of the ID dataset.

\section{Relation between ID classification and OOD detection performance}

In an earlier iteration of the work, we experimented to see if a neural network's in-distribution classification accuracy and OOD detection performance are correlated. This would imply that deploying a neural network with higher in-distribution accuracy will also make it effective at OOD detection.

\begin{figure}
    \centering
    \includegraphics[width=\columnwidth]{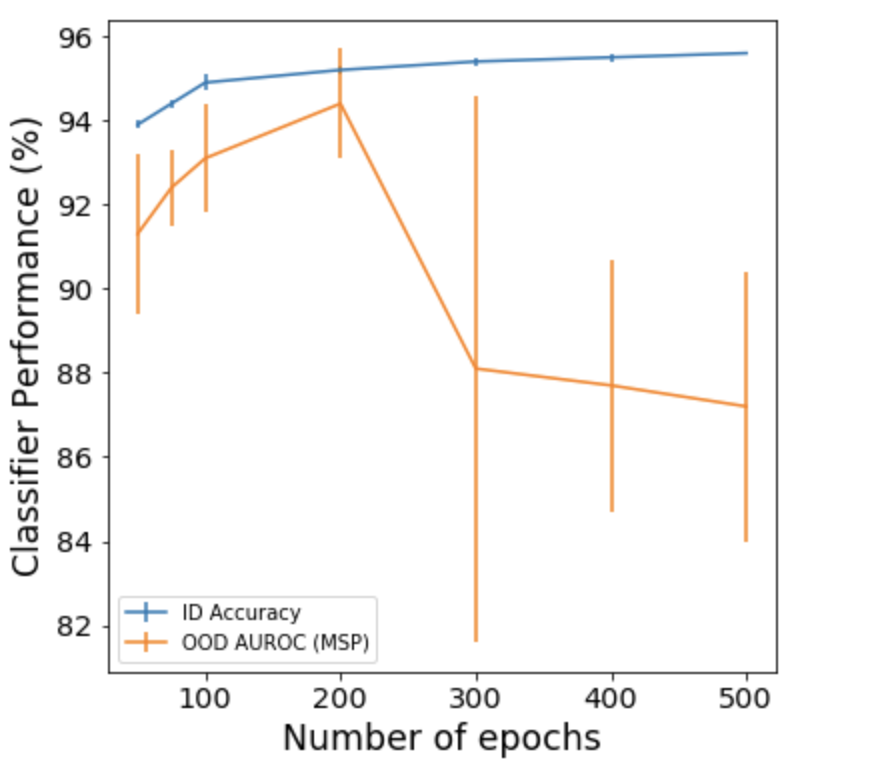}
    \caption{ID classification accuracy and OOD detection AUROC vs number of epochs, for (ID, OOD) = (CIFAR-10, SVHN). For each $N_{epoch} \in \{50, 75, 100, 200, 300, 400, 500\}$, we train three WideResNet-40-2~\cite{zagoruyko_16} networks for $N_{epoch}$ epochs, and plot the average values. The standard deviations are used for error bars.}
    \label{fig:numberOfEpochs}
\end{figure}

However, this turned out to be not true in general. Here we report one such experiment. We train a WideResNet~\cite{zagoruyko_16} architecture with depth 40 and widen factor 2 on the CIFAR-10 dataset for varying number of epochs, similar to~\citet{hendrycks_19}. In particular, we train neural networks for 50, 75, 100, 200, 300, 400 and 500 epochs, with batch size 128, SGD with Nesterov momentum, initial learning rate 0.1 and cosine annealing learning rate~\cite{loshchilov_17}. We notice that the neural network always improves on ID classification accuracy as the number of epochs increase. However, the OOD detection performance for (ID, OOD) = (CIFAR-10, SVHN) seems to drop once the number of epochs increase beyond 200. For a particular number of epochs, we train 3 neural networks and plot the average values in Figure \ref{fig:numberOfEpochs}, with the standard deviations as error bars.

Additionally, Figure \ref{fig:numberOfEpochs} shows that increase in ID classification accuracy does not always translate to an increase in OOD detection performance. Furthermore, we see that the standard deviation on OOD detection AUROC is much larger than the standard deviation for ID classification accuracy --- so model performance on OOD detection task fluctuates more than its performance on ID classification task.

\end{document}